\documentclass[10pt,journal,compsoc]{IEEEtran}
\ifCLASSOPTIONcompsoc
  \usepackage[nocompress]{cite}
\else
  \usepackage{cite}
\fi
\usepackage[justification=centering]{caption}
\ifCLASSINFOpdf
  \usepackage[pdftex]{graphicx}
  \graphicspath{{Figures/}}
\else
\fi
\usepackage{amsmath}
\usepackage{amsfonts}
\usepackage{longtable}
\usepackage{algorithmic}
\begin{document}
%
\title{RGB-Depth SLAM Review}
%
%
%
%


\author
{Redhwan~Jamiruddin, Ali~Osman~Sari, Jahanzaib~Shabbir, and
 Tarique~Anwer}

\markboth{Journal of \LaTeX\ Class Files,~Vol.~14, No.~8, August~2015}%
{Shell \MakeLowercase{\textit{et al.}}: Bare Demo of IEEEtran.cls for Computer Society Journals}
%



\IEEEtitleabstractindextext{%
\begin{abstract}
Simultaneous Localization and Mapping (SLAM) have made the real time dense reconstruction possible increasing the prospects of navigation, tracking, and augmented reality problems. Some breakthroughs have been achieved in this regard during past few decades and more remarkable works are still going on. This paper presents an overview of SLAM approaches that have been developed till now. Kinect Fusion algorithm, its variants and further developed approaches are discussed in detailed. The algorithms and approaches are compared for their effectiveness in tracking and mapping based on Root Mean Square error over online available datasets.
\end{abstract}

}

\maketitle

\IEEEdisplaynontitleabstractindextext

%
\IEEEpeerreviewmaketitle

\IEEEraisesectionheading{\section{Introduction}\label{sec:introduction}}

%
%
%
%

 

\section{Taxonomy of different methods}

Simultaneous Localization and Mapping (SLAM) have made the real time dense reconstruction possible increasing the prospects of robot navigation, tracking, and augmented reality problems . Some breakthroughs have been achieved in this regard during past few decades and more remarkable works are still going on. This paper presents an overview of SLAM approaches that have been developed till now. Kinect Fusion algorithm, its variants and further developed approaches are discussed in detailed. The algorithms and approaches are compared for their effectiveness in tracking and mapping based on Root Mean Square error over online available datasets.
\subsection{RGB-Depth Mapping}
As far as an optimal perception of phenomenal consciousness is concerned, theories based on representation of the mind are based on models of the information processing paradigm \cite{turan2018unsupervised}. These are as much in correspondence to the neurobiological or functional theories, at this point we are confronted with several arguments on the basis of inversion or absent qualia \cite{altuntacs20163}. 
Such considerations exhibit a preceding pattern based on the assumption of holding complete knowledge of the neural and functional states that are in subservience to the occurrence of the consciousness that is phenomenal. This can still be conceived as the neural states which are also defined as the states with similar casual responsibilities or with similar representational function \cite{turan2017deep1}, \cite{turan2017endosensorfusion}.

\subsection{Spatially Extended and Moving Volume Kinetic Fusion}
These occur with no phenomenal content in any way or such states being accompanied by contents that are phenomenal with broad variation from the usual ones. In definition, visual information processing entails the visual cognitive skills that permit us the processing and interpretation of meaning from visualized information that we attain through eye sight. 
Therefore, visual perception plays are vital role in aspects of cognitive and intelligence skills such as spelling, math and reading (\cite{angonese2016integration}). On the other hand, visual perceptual deficits can lead to challenges in learning, recognition and remembrance of letters, wording, and confusion of likeness as well as minor variations in addition to differentiating the main ideal from the details of insignificance.

\subsection{Scalable Real-Time Volumetric Reconstruction}
Visual perceptual processing can be sub segmented into the categories that comprise of visual discrimination, figure grounding, closure, memory, sequential memorization, constancy, spatial relations as well as visual motor integration. 
Note should be taken of perception as active procedures of location and information extraction form the setting while learning entails the procedures of acquisition  of information through experiences of information storage. In which case, thought is the manipulative stance upon information for solving challenges (\cite{ataer2016object}). 
Such that it is eased to extract information (perception) which creates an ease in thought procedures becoming. In overall it is accepted that human vision takes the form of extreme powerful processing of information towards facilitation of the interaction of the world that surrounds us. 
However, even in the face of extended and extensive efforts of research encompassing multiple fields of exploration, the fundamentals that underlay as well as operational principles of visual information procedures remain largely unknown.

\subsection{Segmentation-Based RGB-D Mapping}
We are still not able to ascertain the origin and distance along the route from eyes to the sensory input area known as the cortex. It is in this area that the conversion into object meaningful representation is undertaken under conscious manipulation of the brain (\cite{ataer2013tracking}).  
Nearly half of the human brain in the cerebral cortex region is charged with the processes of visual information although even with extended and extensive research efforts that are encompassed a conundrum still persists. Present theories on visual information processing are held in the consideration of human visual information processing being interplay of the two inversely directed procedural streams. 

\subsection{B-D Visual Odometry}
This is taking the form of a non-supervised, top down directed procedures that convey the regulations and guiding knowledge as a guide to linkage and binding of disjointed pieces of information to meaningful perceptual object images (\cite{altuntacs20163}). Most important in the idea of such a proposition not completely being new as in past research, there have been presentations in the form of depictions of the "faculty of appreciation" as a synthesized relation of "two constituents" which include the raw sensory data with the other being the cognitive "faculty of reason."

\subsection{Elastic Fusion}
Past research has presented a demonstration of distance and physical enviroment being among the aspects that impairs processing of information, although it remains unknown whether such impairment is on all the levels of information processing or in the onset states instead of the later stages. 
Those faced with the condition of mapping algorithms suffer from deficiencies of attention that are impairment to the capability of selective procedures of visual information that is incoming. The early levels of information processing are held in the description of being those that entail the detection as well as response of simplified stimuli.
 An assignment on the assessment of such function is the inspection time that has previously been demonstrated to entail sensitivity to pharmacological agents. This is as well as being the most reliable and validated within the cultural fairness of information processing measures of cognitive ability (\cite{angonese2016integration}). 
Past assessment findings have also presented the impact of nicotine on information procedures as being held in the overall regard in the form of a measure of speed within the early levels of information processing. These include the speed of visual encoding that comprises of the ability of making observations or inspections on sensory input on which the discrimination of relative magnitude rests. 
This is in contrast to assignments such as reaction time which is summarization entails the involvement of increased response oriented measures of complete decision making time that comprise of total information processing. 
Although, there is no research of examination of the impacts administration of 3D scene construction in a similar response, there are limited studies based on the examination of the impacts of 3D scene construction in the early stages of information processing with utilization of other assignments (\cite{ataer2016object}). With the application of visual tracking assignments, it was ascertained that the speed of detection experienced impairment from 3D scene construction that that these impacts where greater in dual task settings with comparison to single task settings. 
Such outcomes have been held in the description of being the deleterious impacts of 3D scene construction on the centralized processing capacity and on information processing availability on the capacity of information processing with time. Further investigations of early information processing are based on the examination of the mismatched negative component of auditory event relation potential as well as reports of reduced dosage of 3D scene construction attenuation of the event relation potential signal. 
In this case, the mismatched negative component suppression was solid within stimuli deviation as reduced which the indication of relatively reduced blood 3D scene construction concentration is.  The detection of minimal deviations for instance that needed in the course of the inspection time assignment more so in case of hampering in which case similar outcomes have been discovered in simplified reaction time assignments with double level of intensified stimuli. 
These studies produced outcomes of an increase in response time as well as the impairment of stimuli detection which is a suggestion of the influence on sensory perceptual procedures and the measure of attentiveness (\cite{ataer2013tracking}).

\subsection{Bundle Fusion}
Current discoveries in the arena of visual information processing are based on the reflection of the elementary principles of vision as well as the utilization of visual information based on cognitive attributes. 
This is based on the notion of such work leading to the verge of development based on the grounds of optimism within the several computational theories of sophistication that incorporate data that is neurobiological and behavioral. 
These theories entail the flourishing of the skillful exploitation of the neural-imaging and computation of simulative technologies, these permits answering of questions that are subtle regarding the component subsystems within vision. 

\section{Most relevant methods and description of their novelties and contributions and why are they published}

\subsection{GRAPH SLAM}
This algorithm applies information matrices sparsely production by the generation of graphs using observed interdependencies in case the observations are connected and if they contain information about the similar landmark.
Graph SLAM allows for the capability of constructing a map from an environment while simultaneously creating associated localization with the map for navigation in unknown settings when external referencing systems such as GPS are absent.  This intuitive approach utilizes a graph with nodes in correspondence to the robot poses at varied points within time and whose edges are representative of the constraint in between the poses. 
The latter is gained from environment observations of from movement actions as performed by the robot. Upon construction of a graph, the map could be computed by searching the nodes spatial configuration that is notably consistent with modeled measurements by the edges. 

\begin{figure*}
	\centering
	\includegraphics{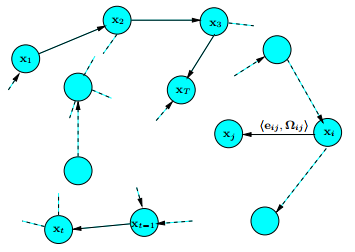}
	\caption{Graph-pose SLAM estimation procedure}
\end{figure*}

From the image above, we note that particular nodes within the graph are in correspondence to the pose of the robot. Proximal poses are linked by the edges with model spatial constraints between the robot poses that are derived from measurements among the consecutive poses of model odometry measurements. This is whereas the other edges are representative of the spatial constraints based from several observations of the similar section of the environment.
The graph-based SLAM method develops a simple estimation challenge by abstraction of raw sensor readings. These readings as substituted by the graph edges which are viewed as "virtual measurements". Increased detail within an edge between the two nodes holds the label of a probability distribution over locations that are relative to the two poses with conditioning to mutual measurements. 

\subsection{RGB-D Camera-Based Parallel Tracking and Meshing}
Visual real-time tracking in regard to established and unknown scenes is critical as well as an incontrovertible aspect in vision-based AR applications. Multiple algorithm contributions over the years. It is at this point that we introduce RGB-D Camera-Based Parallel Tracking and Meshing as an adaptation and updating of the algorithms utilized in estimating the motion of the camera as well as AR in accordance to the availability of the end user in computational abilities in permitting to gain impressive tracking outcomes in limited AR workspaces. 
The fact is that estimation of camera motion using environment tracking as well as parallel constructing feature based sparse mapping that creates a possibility in part to the generalization of multi-core processors found in desktop and laptop computers. Of recent is has been revealed that increased computation power within a singular standard of a hand-held video camera is connected to a powerful computer using computational power gained from the Graphics Processing Unit (GPU). 
The possibility to attain a dense representation of a desktop setting as well as increased texturing scenery whereas as undertaking tracking with the use RGB-D Camera-Based Parallel Tracking and Meshing. The online created map density can be increased with the use stereo-dense matching in addition to GPU founded implementations as shown by GPU to be utilized for effective replacement of the global bundle adjustment aspects of SLAM optimized based systems for instance RGB-D Camera-Based Parallel Tracking and Meshing as well as inherent parallelization refinement with step founded Monte Carlo simulations therefore freeing tools on the CPU for other assignments.

\subsection{LSD-SLAM (Large Scale Direct Monocular SLAM)}
The following algorithm is a novel direct monocular SLAM method that operates with direct image intensities rather than the use of key points for tracking and mapping. Camera tracking utilizes direct image alignment whereas geometry is estimated in the format of semi-dense depth maps gained through filtration of several pixel wise stereo comparisons. Thereafter, a Sim (3) pose-graph of key frames is created to permit for the development of scale-drift correction with large scale maps comprising of loop-closures. It should be noted that LSD-SLAM could be operated in real time on a CPU as well as smartphones.
This algorithm comprises of three core components namely tracking, map optimization and depth map estimation. The tracking feature persistently tracks new camera images by estimating the rigid body pose in regard to the present key frame and the uses of the pose in the past frame as a point of initialization. On the other hand, the depth map estimation feature applies tracked frames for either refinement or replacement of the present key frame. 
Refinement of depth is achieved by filtration over several per-pixel, limited based line stereo comparisons as well as interleaved spatial regularization as the default proposition. Should the camera extend to far, initialization of a new key frame is implemented by projection of points from existent and proximal key frames. 
Furthermore, upon replacement of a key frame as a reference tracking, the depth map will not be additionally refined but rather integrated into the global mapping with use of the map optimization feature. In this case, for detection of loop closures as well as scale drifting, the same transform to proximal key frames inclusive of the direct predecessor is estimated with use of the scale-aware and direct image alignment.

\subsection{S-PTAM (Stereo Parallel Tracking and Mapping)}
The algorithm holds the capability of computation of camera trajectory in real time with heavily exploitation of the parallel format of the SLAM challenge, separation of time constraints in pose estimation from less pressing issues for instance building of maps and refinement of assignments. In addition, the stereo setting permits for reconstruction of a metric 3D map for particular stereo frame d images, improvement of mapping procedure accuracy in respect to monocular SLAM and limiting the common bootstrapping challenge. Furthermore, the actual scale of the environment is a critical aspect for robots when in it comes to interaction with the surrounding workspace.
In order to permit for robotic mobile navigation and achieve autonomous assignments, it must be understood for its pose (position and orientation) as well as hold an environment (map) representation. In settings where robots do not have a past map and external information availed of the pose, it is necessitated to undertake both assignments simultaneously. The challenge of the robot and constraint the map of the environment in a simultaneous action is known as SLAM. However, in order to tackle the challenge of stereo vision, we introduced the S-PTAM (Stereo Parallel Tracking and Mapping) algorithm as an approach whose intention is to operate real-time of an extended duration of lengthy trajectories to permit for estimation of the pose with accuracy as it is built upon sparse mapped environments with a global coordinate system \cite{turan2017deep},\cite{turan2017endo}.
By using optimal performance, this algorithm is able to decouple localization and mapping assignments for the SLAM challenge with two independent threads which permits us to take the benefit of multi core processors. In addition to localization as well as mapping modules, the loop closure function is able to recognize locations from historically visited points. These detected loops are then applied for refinement of the map and trajectory estimation to effectively lower the accumulated error of the method. It is on this basis that S-PTAM operates on the visual features from extraction of images availed by the stereo camera.

\section{3D scene reconstruction and mapping}
3D scene reconstruction and mapping has been a crucial and important assignment within the arena of moveable robotics since it is critical need for various techniques specifically including path planning, semantic mapping, localization, navigation, and telepresence. Two major approaches towards 3D reconstruction are: offline multi-view stereo (MVS) based reconstruction and live incremental dense scene reconstruction. Many compelling results have been produced since past few years by exploring multi view stereo (MVS) and format derived on the basis of motion (SfM) techniques. Multi perspective stereo has been used extensively in photogrammetry for dense surface reconstruction (\cite{seitz2006comparison}) while the problem of accurate camera tracking has been cattered by SfM algorithms along with sparse reconstruction from large datasets of unordered images (\cite{agarwal2010reconstructing}). Although some groundbreaking results have been achieved but most of both SfM and MVS approaches have not been driven by live implementations. 
Simultaneous Localization and Mapping (SLAM), unlike SfM and SVM, provides live motion tracking and re-structuring while applying input from a single commodity sensor but for a sequential ordered set of images. Various 3D mapping techniques offer different functionalities but all of them work almost on the same pipeline; of spatial aligning consecutive data frames at first, detecting the loop closures, and aligning the complete data sequence in a globally consistent manner. Although the developed systems provided satisfactory accuracy through point clouds and colored cameras but most of them are computationally exhaustive and inaccurate for dense depth reconstruction especially in dark environments or scenes with sparsely textured features \cite{turan2018magnetic}.
Based on the sensors used, 3D reconstruction can be achieved via three routes: using Multiview stereo, Laser scanning, or depth cameras. Multiview stereo is the traditional technique of photogrammetry where overlapping multi views of an object are captured for relative camera pose estimation and scene reconstruction is done via selected control points to get 3D coordinates of the object’s points through space intersection. Laser scanners work on the principle of time of flight where scene tracking is achieved via transmitted laser pulses that are received back by the scanner with high accuracy. The most recent and popular approaches are of constructing 3D scenes using RGB depth cameras that, working on the principle of time of flight, measure the pixel depth along with color information of the pixels. 
Some early work on SLAM in 3D reconstruction over past few decades includes a range of approaches and their extensions. 3D reconstruction has been explored extensively with some point cloud models with real-time tracking like MonoSLAM (\cite{davison2007monoslam}) being the first successful effort on real-time tracking and active 3D mapping with only one camera. This had motivated many other works for online, though sparse, but fine and accurate reconstruction with freely moving hand-held cameras based on probabilistic models (\cite{weise2009hand}). Some later research focused on performing tracking and mapping in parallel instead of adopting probabilistic models. Parallel Tracking and Mapping system (PTAM) worked on the hierarchy of live tracking via feature optimization over spatially-distributed key frames for n-point pose estimation and expanding the maps obtained via bundle adjustment and global pose optimization (\cite{klein2007parallel}). 
Although the mono SLAM approaches set the benchmarks in real-time 3D mapping and developed robust camera tracking systems, but the AR (Augmented Reality), and other live robust mapping and robot navigation applications cannot rely on sparse point clouds generated as a result of these systems. This triggered the work towards generating live dense maps using depth information of the scene via Multiview stereo approaches combined with PTAM for live camera tracking and robust pose estimation (\cite{newcombe2011kinectfusion}). But the availability of depth camera has made the task further easier and current approaches have set their focus on large scale 3D mapping using depth commodity sensors. 
Considering the importance of SLAM approaches and their applications in field of robotics, this paper reveals a general understanding of the development of SLAM approaches for dense surface mapping and reconstruction in real-time using depth cameras as commodity sensors. An introduction of Kinect sensor is presented with its unique use in depth mapping and reconstruction for Augmented Reality (AR) applications. The focus is set on KinectFusion algorithms and marks achieved from them or their integration with other tracking and mapping algorithms. \cite{DBLP:journals/corr/abs-1804-01396}

\section{Depth Surface Mapping and Tracking Algorithms for 3D Reconstruction}
\subsection{RGB-Depth SLAM}
Depth cameras, with their ability to measure object’s depth from camera (based on time-of-flight or active stereo) in addition to RGB measurement, have paved a new wave of techniques in SLAM and Augmented Reality (AR). Incorporation of RGB-D cameras has allowed SLAM to benefit from range sensing along with visual data to handle the issues like data association and loop closures in visual Odometry along with visual SLAM (\cite{agarwal2010reconstructing}). Kinect sensor, among other RBG-D cameras, is the most notable depth device to be used in revolutionary approaches being developed for real-time tracking and surface mapping algorithms.

\subsection{Kinect Sensor}
Kinect sensor, a low-cost commodity platform mainly to detect human gestures in gaming and other entertainment applications, has shown its potential in simultaneous localization and mapping approaches to an unprecedented level. It applies an internal ASIC to generate 11-bit 640x480 depth map of a pixel at 30 Hz. Although map quality suffers from certain technical challenges (like motion blur at faster speeds), the information available is significant enough to be utilized by real-time 3D reconstruction algorithms. There have also been algorithms available to improve sensor accuracy (\cite{karan2015calibration}, \cite{butler2012shake}) depending upon sensor’s use or system’s requirements. 

\subsection{Kinetic Fusion}
Developed by \cite{newcombe2011kinectfusion}, KinectFusion algorithm was the first attempt to real-time volumetric reconstruction of a scene in variable lightning conditions (\cite{newcombe2011kinectfusion}). Using information gained through Kinect sensor in form of input, while utilizing a coarse-to-fine iterative closest point (ICP) algorithm to simultaneously track camera pose and construct a medium sized 3D model in real-time by tracking a live depth frame relative to a global finished model. At a given time k, the transformation matrix given below was used to describe the 6 DOF, that mapped the camera coordinate frame to a global frame g, such as shown in \ref{eq1}.

\begin{equation}
\label{eq1}
 T_{g,k} = \begin{bmatrix} R_{g,k} & t_{g,k} \\ 0 & 1 \end{bmatrix} \in \mathbb{SE}_3
\end{equation}

\begin{figure*}

	\centering

	\includegraphics[width=6in,height=1.5in]{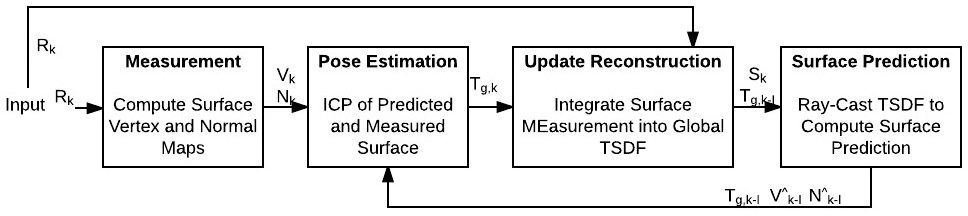}
	\caption{KineticFusion Workflow for real-time volumetric reconstruction}
\label{fig2}
\end{figure*}

In equation \ref{eq1}, $\mathbb{SE}_3 := \{R,t|R \in \mathbb{SO}_3, t \in\mathbb{ R}^3\}$. This means, any point $P_k \in\mathbb{ R}^3$ in the camera frame is mapped to global coordinate frame via transformation $P_g=T_{g,k} P_k$. The algorithm was able to do real-time volumetric reconstruction in four steps – surface measurement, surface reconstruction update, surface prediction, and sensor pose estimation (Figure \ref{fig2}) – explained below: 

\textit{Surface Measurement}: The first step is Pre-ICP registration and subsampling where raw depth measurements from Kinect sensor are used to build depth map pyramids (normal map and vertex map pyramid). A depth map $R_k (u) \in \mathbb{ R}$ obtained at every pixel $\bold{u} = (u,v)^T$  is calculated to obtain a metric point measurement $R_k (\bold{u}) K^{-1} \dot{\bold{u}}$ in the sensor frame at time k. A depth map $D_k$ is obtained after applying the bilateral filter (\cite{newcombe2011kinectfusion}), which after back projecting the value obtained into the sensor frame of reference gives the vertex map $V_k (\bold{u}) = D_k (\bold{u}) K^{-1} \dot{\bold{u}}$. Further, applying the rigid body transformation given in equation \ref{eq1}, the global frame vertex is obtained as: $V{_k^g} (\bold{u}) = T_{g,k} \dot{V}_k (\bold{u})$ while the equivalent mapping of normal vectors in the global frame is given as: $V{_k^g} (\bold{u}) = R_{g,k} N_k (\bold{u})$. \\
\textit{Surface Reconstruction Update}: a truncated signed distance function (TSDF) is computed from each input sensor frame. At each location, the global TSDF $S_k (\bold{p})$ at a given global point $\bold{p} \in \mathbb{ R}^3$ is a combination of local TSDF $F_k (\bold{p})$ and a weight  $W_k (\bold{p})$ such that: $S_k (\bold{p}) \rightarrow [ F_k (\bold{p}) , W_k (\bold{p}) ]$. A volumetric integration fuses this TSDF into a discretized volumetric representation of the global coordinate space.\\
\textit{Surface Prediction from Ray-casting the TSDF}:  It is a post-volumetric integration step in which the surface is visualized by ray-casting the signed distance function into an estimated frame and aligning this ray-casted view with live depth map. In pixel vise ray casting each ray $T_{g,k} K^{-1} \dot{\bold{u}}$ is started from a zero depth value till the surface crossing is found. \\

\begin{figure*}
	\centering
	\includegraphics[width=6in,height=3in]{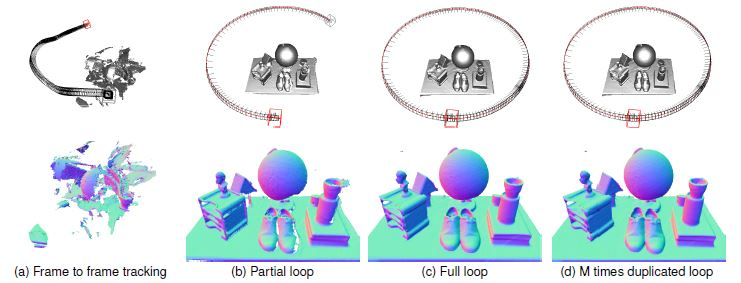}
	\caption{Reconstruction results by Kinect fusion system through normal mapping (constant reconstruction resolution of 2563 voxels) with different sensor trajectories – Courtesy: Newcombe et al (2011)}
\label{fig3}
\end{figure*}

\textit{Sensor Pose Estimation}: a multiscale ICP alignment is done between current sensor measurement and predicted surface to achieve sensor tracking. The algorithm is implemented on GPU to utilize all the data available at frame-rate. 
The system shows notable accuracy for metrically consistent reconstruction from local loop closure trajectories without performing explicit global joint optimization (Figure \ref{fig3}). The system shows its effectiveness in reducing drift too by performing tracking relative to key frames instead of frame-to-frame matching. System can be easily scaled to available GPU memory with slightly lesser resolution.

Tracking drifting occurs when sensor is faced with large planner scenes which accounts for system’s shortcomings, but Kinect fusion provides a powerful basis for large scale volumetric reconstruction and dense modeling with various approaches projected by \cite{newcombe2011kinectfusion}.
The Point Cloud Library developed by Rusu and Cousins \cite{rusu20113d} implements the Kinect fusion algorithm to develop Kinfu: an open source implementation hirearchy along with other methods for point clouds manipulation and 3D reconstruction. 
Another extension of Kinfu is developed recently by Korn and Pauli with an alternative algorithm for ICP for increased voxel grid hence improving scene dynamics scaning \cite{korn2015kinfu}. The voxel grid data used by Kinfu is used to create vertex and normal maps that are registered with the maps obtained from sensor. But in doing so, unusual amount of information is lost. To cater this problem, Korn and Pauli have suggested direct matching of the maps obtained from sensor with voxel grid model. The ICP algorithm developed by them is also different from the original ICP algorithm adopted for Kinfu as they,ve removed the normal threshold and use the normals computed from the depth maps for point-to-plane error metric instead of using normals from voxel grid that has shown improved robustness in terms of pose estimations with moving objects. 

\subsection{RGB-Depth Mapping}
\begin{figure*}
	\centering
	\includegraphics[width=6in,height=2in]{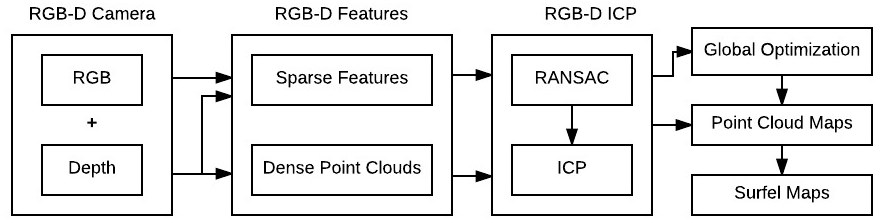}
	\caption{Workflow for RGB-D mapping for constructing 3D models of indoor scenes}
\label{fig4}
\end{figure*}

As kinetic fusion algorithm provides consistent and accurate volumetric reconstruction of smaller indoor scenes, the problem of dense 3D mapping of large indoor environments is addressed by the RGB-Depth mapping algorithm by \cite{henry2012rgb}, a framework that uses RGB depth camera to generate dense 3D models of even darker and featureless planner indoor environments. A joint optimization computed over object shape as well as appearance matching (RGB features) is computed to develop alighnment between the frames followed bysparse features extraction and matching using RANSAC. Loop colosures are detected via matching data frames compared to a subset of earlier collected frames and finally an improved, globally consistent allignment is completed either via sparse bundle adjustment (SBA) or a more efficient pose grapgh optimization that is TORO in this case. 
What lies at the core of RGB-D mapping is its novel ICP algorithm \cite{turan2017fully}, RGB-D ICP (Figure \ref{fig4}), that identifies the sparse feature points in each camera frame using the visual information. These identified point features then help in RANSAC optimization. An RGB-D frame $P_s$ is input to RGB-ICP algorithm along with target frame $P_t$. For an instant’s rotation R and translation t, the rigid transform is $T(p)=R_p+t$. RANSAC then finds the best optimized rigid transform $T^*$ in order to get best alignment as shown in Equation\ref{eq2}.

\begin{equation}
\label{eq2}
T^* = argminT\Bigg(\dfrac{1}{|A_f|} \Sigma_{i \in A_f} |Proj(T(f_s^i)) - Proj(f_t^i)|^2 \Bigg)
\end{equation}

In Equation\ref{eq2}, $A_f$ is the set containing correspondences between the features points of the target frame $f_t^i$ and source frame $f_s^i$ and the projection function finds the projection of feature points $(x,y,z) \in \mathbb{ R}^3$ in Euclidian space to feature points in camera space $(u,v,d) \in \mathbb{ R}^3$ where d is the depth of each pixel $(u,v)$. A joint optimization performed after that minimizes the re-projection error resulting in an even refine alignment. The joint optimization function is actually modeled to minimize the alignment error of both visual feature association and depth point association and is given as shown in Equation \ref{eq3}.

\begin{multline}
T^* = argminT\Bigg[\alpha\Bigg(\dfrac{1}{|A_f|} \Sigma_{i \in A_f} w_i |T(f_s^i) - f_t^i|^2 \Bigg) + \\
(1-\alpha)\Bigg(\dfrac{1}{|A_d|} \Sigma_{j \in A_d} w_i |\big(T(P_s^j) - P_t^j\big).n_t^j|^2 \Bigg)\Bigg]
\label{eq3}
\end{multline}

As shown in Figure \ref{fig4}, the system has a two-staged RGB-D ICP that combines the best properties of both sparse visual features and dense point clouds to get accurate and robust alignment in case of visual features or in complete darkness to determined accuracy. The features are detected using FAST feature detector combined with the Calonder feature descriptor to obtain even better results than using SIFT feature detector as used in the previous work of this study \cite{henry2010rgb}.
Surfel mapping (mapping via small colored surface patches) allows an efficient estimation of surface orientation, color extraction and visualization of the model \cite{krainin2011manipulator}. \cite{henry2012rgb} have proposed window-based sequential bundle adjustment to increase alignmnet accuracy of the system, while loop closure detection is sugested to improve by considering all the depth and color information instead of frame-to-frame visual matching. 

\subsection{Spatially Extended and Moving Volume Kinetic Fusion}
As ICP based kinetic fusion suffers from the limitations of bounded reconstruction with poor tracking results for planner geometries. An alternative to ICP is Fast Odometry from Vision Systems (FOVIS) system by \cite{huang2017visual}, a framework based on standard stereo odometry pipeline that reduces mismatches and hence overall drift and increases robustness \cite{huang2017visual}. \cite{whelan2012robust} have updated the algorithm with a combination of ICP and FOVIS to cater for the two challenges stated above that generates a continuous fused map of a scene at an unprecedented extended scale through triangular mesh generation in real time. Unlike \cite{newcombe2011kinectfusion}, the focus of the kintinuous algorithm is to allow the area mapped by the TSDF to move over time to keep the origin of the TSDF at the center of TSDF volume. The movement threshold b is the distance in meters in all the directions from the current position of TSDF origin before re-centering of TSDF. Once the threshold is crossed (referred as boundary crossing in \cite{huang2017visual} given as $v_m = \dfrac{d}{v_s}$), the new camera pose is calculated as shown in Equation \ref{eq4}.

\begin{equation}
C_{i+1} = (R_{i+1}, t'_{i+1})
\label{eq4}
\end{equation}

\begin{figure*}
	\centering
	\includegraphics[width=6in,height=1.8in]{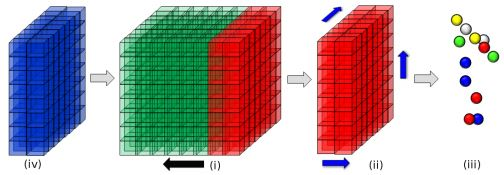}
	\caption{Extended Kintinuous pipeline for robust camera tracking and real-time colored dense mapping (i) vertex and normal maps are computed from Kinect fusion using the raw depth maps (ii) using a weighted running average (ray casting), the depth data is integrated into current TSDF resulting in a smooth surface reconstruction (iii) point cloud is extracted and triangulated to get mesh surface (iv) new region is registered to volume}
\label{fig5}
\end{figure*}

In Equation \ref{eq4} $C_i$ and $t_i$ are the 6 DOF camera pose and translation vector at time i. And the new position of the global position is calculated by adding the number of number of whole voxel units crossed since the last boundary crossing $\bold{u}=[\dfrac{t_{i+1}}{v_m}]$ to the current global position of TSDF such that: $g_{i+1}=g_{i+\bold{u}}$.
In extension of their work in [16], Whelan et al present the complete workflow of GPU based kintinuous algorithms with a combination of ICP and FOVIS to generate high quality dense colored maps in real-time with robustness (Figure \ref{fig5}). They have adopted a novel switching strategy for which the system can switch between FOVIS and some other estimator depending on the error metric such that the system is accustomed to use FOVIS camera transform $T_F$ when $\bigg|\Big|\big|t_F\big|\Big|_2-\Big|\big|t_O\big|\Big|_2\bigg|> \mu$ otherwise, the system uses other estimator’s transform $T_O$ which is ICP for their system. $\mu$ Here is an experimentally set error metric which was kept 0.03m for system discussed in \cite{whelan2012kintinuous}. The chosen transform is then used to calculate next camera pose. The color and depth information is then combined in RGB-D and ICP integration step where two functions are weighted and combined in one cost function defined as: $E = E_{icp}+ \omega.E_{rgbd}$, where $\omega$ is the assigned weight on empirical calculation basis. 
The voxels are streamed out of GPU based camera motion allowing space for new set of data. The streamed voxels are compressed to greedy mesh assuring a fine-quality reconstruction. Instead of restricting tracking and reconstruction to TSDF initialization points, Kintinuous has introduced a cyclic buffer type data structure that allows the area mapped by TSDF to move over time that results in an augments reconstruction of scene incrementally with time. A parallel processing system with multi-level threaded algorithms enables a continuous TSDF tracking and reconstruction of extended areas with unparalleled accuracy. 

Rothe and Vona propose the same approach in \cite{roth2012moving} but for a moving volume, i.e, volume is also rotated while translated making the algorithms adaptable to free-roaming applications. Using a 6DOF visual odometry approach, at any time b the local camera pose ${_b^a}C$ is obtained relative to an earlier time a. Based on a threshold value it is determined if a new volume frame is needed or not. In case of a new volume frame, the new frame is remapped from the old by swapping the TSDF buffer and the one in the GPU memory such that new volume transform $P_{n+1}=C_t C_{t+1}^{t-1}$ is set. The idea is to translate and rotate the TSDF volumes that results in 6DOF visual odometry with camera poses constant relative to scene via volume-to-volume transforms hence generating an always-available 3D map of the scene. The linear $l_d$ and angular $a_d$ camera offsets are calculated using equation \ref{eq5} to obtain a local camera pose $C_t$ that determines whether there is a need of a new volume frame or not. 

\begin{multline}
D = \begin{bmatrix} R_d & t_d \\ 0 & 1 \end{bmatrix} = C_s^{-1} C_t,
C_t =  \begin{bmatrix} R_t & t_t \\ 0 & 1 \end{bmatrix},\\
l_d = \Big|\big| t_d \big|\Big|,
a_d = ||rodrigues^{-1}(R_d)||
\label{eq4}
\end{multline}

Some other approaches have also focused on multi-resolution RGB-D Scene Mapping. One such approach is Omnikinect (\cite{kainz2012omnikinect}) that offers a physical multiple-Kinect infrastructure along with tracking, filtering, and visual hull rendering software for real-time data acquisition and volumetric reconstruction at different resolutions. They have modified the Kinect fusion pipeline in a way that the new volumes are introduced as discrete histograms for the TSDF per voxel such that each voxel $v$ for a deviced, the TSDF is a function $f_{R_k} (v,d)$ for a distance $\eta$ to a measured depth is evaluated for the values $\Psi(\eta)$ in one histogram $\bold{\theta}(v)$ per voxel with an increment in the corresponding histogram bin with $\lambda$. After the TSDF for each value is evaluated the histograms are filtered separately giving filtered TSDF with reduced noise that improves the calibration accuracy of the pose estimation step of original Kinect fusion algorithm.

\subsection{Scalable Real-Time Volumetric Reconstruction}
Although algorithms like kintinuous and moving volume Kinect fusion allow more space for storing data by streaming voxels in real-time, the size of the active volume remains restricted which has triggered the research for saleable volumetric dense mapping and reconstruction. Two notable approaches for this are hierarchal scalable reconstruction by \cite{chen2013scalable} and direct volumetric reconstruction by \cite{niessner2013real}. 

\begin{figure*}
	\centering
	\includegraphics[width=6in,height=2.3in]{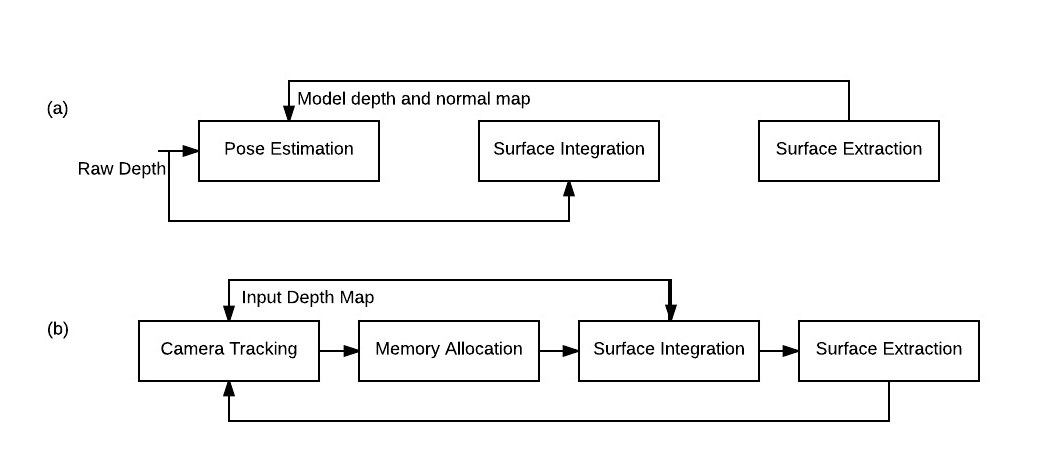}
	\caption{Volumetric reconstruction pipeline through Voxel Hashing}
\label{fig6}
\end{figure*}

Both works have developed an upgraded same volumetric method developed by \cite{curless1996volumetric} to generate scalable volumetric reconstruction in real-time. Chen et al adopts the standard Kinect fusion algorithm (Figure \ref{fig6}-a) while dynamically fusing and updating live depth maps and streaming the generated maps between the GPU structure and host all in a hierarchal manner. The basic idea of dense reconstruction through voxel hashing is to fuse an implicit SDF in the dense data structure avoiding a hierarchal data structure and incorporate a hashing scheme instead to efficiently store, access and update the scene reconstruction implicitly. Chen et al have adopted a GPU implementation of the system using CUDA where each hierarchical grid is saved as a sparse point structure in the GPU memory. The root level grid is a dense 3D array of GridDesc records which are initialized to null. Parallel reduction is performed to keep grid descriptors to nearSurface while minWeight field is used as a heuristic for garbage collection. With previous and current SDF $d_p$  and $d_c$ obtained, it is tested if zero crossing is obtained or not. If the surface is obtained then $t_z = t_p + \dfrac{d_p}{d_p - d_c}$; otherwise $d_c$ is set equal to $d_p$ and iteration is performed again. The gradient of the SDF at zero crossing gives the surface normal. 
Figure \ref{fig6}-b illustrates the general pipeline of the voxel hashing system; provided a new depth map, the system allocates occupied voxels into hashed blocks (in surface integration) making an implicit surface that is then ray casted for camera pose estimation. \cite{niessner2013real} have proposed a hash table to allocate and retrieve the voxel blocks where the following hashing function maps the world coordinates $(x,y,z)$ to hash values $H(x,y,z)$ such that as shown in Equation \ref{eq5}.

\begin{equation}
H(x,y,z) = (x.p_1 \oplus y.p_2 \oplus z.p_3) mod n
\label{eq5}
\end{equation}

In Equation \ref{eq5} $p_1, p_2, p_3$ are very large prime numbers and n is the size of the hash table. The estimation is done frame-to-model by point-plane ICP hence alleviating any drift. In the final step the algorithm performs lossless bilateral streaming (by revisiting previously scanned areas) between the GPU and user. Streaming alleviates the restriction of range and scan revisiting unlike hierarchal approaches of \cite{whelan2012kintinuous} and \cite{chen2013scalable}. 

\subsection{Segmentation-Based RGB-D Mapping }
\begin{figure*}
	\centering
	\includegraphics{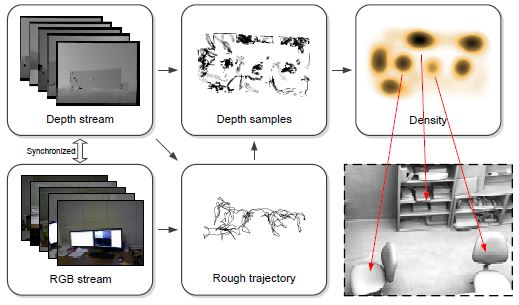}
	\caption{Point of interest detection pipeline: the strong modes in the density function represent points of interest}
\label{fig7}
\end{figure*}

Another rout to fine quality and globally consistent volumetric reconstruction is through segmentation based RGB-D mapping. 3D mapping with points of interest (POI) \cite{zhou2013dense} and with patch volumes \cite{henry2013patch} are two distinguished approaches developed in towards this stream. \cite{zhou2013dense} present the pipeline to construct scene geometry by detecting points of interest for a globally consistent mapping (Figure \ref{fig7}).  For each planar set of localized points $\bar{P} = \{\bar{P}_i^k\} = \{\mathbb{P}_g (\tilde{T}_k p_i^k)\}$ the locations of points of interests POI are found by finding the modes in the density function induced by $\bar{P}$ such that as shown in Equation \ref{eq6}.

\begin{equation}
\rho(x) = \Sigma_{i} w_i^k K (\big|\Big|x-\bar{P}_i^k\Big|\big|/h)
\label{eq6}
\end{equation}
In Equation \ref{eq6}, $w_i^k = \tau exp \bigg(-\dfrac{d_i^{k^2}}{2\sigma^2}\bigg)$ is the weight assigned to a point $p_i^k$ and K is Epanechnikov kernel.

An offline optimization framework is incorporated with frame-to-model registration to detect loop closures. After detecting points of interest the framework performs trajectory segmentation followed by a two-pass registration in which for each frame k, a projective SDF $f_k (x)$ is calculated with the assigned weight function $w_k (x)$ for a local range of images $T_k D_k$. Camera pose estimation and global optimization done in this manner generates high quality dense 3D maps. 
\cite{henry2013patch} propose an algorithm focusing on regions of interest – dense patches of space - instead of points of interest as a modification to general Kinect fusion algorithm. Previously developed volumetric fusion approaches suffer from a lack of scalability especially for larger scenes because of larger memory requirement for dense maps. Henry et al have addressed this problem by dividing the scene geometry into patches of volumes that can be replaced in the GPU memory. A relative corrected pose estimate is obtained by the Equation \ref{eq7}.

\begin{equation}
T^* = argminT\Sigma \Big|\big|\varepsilon\big|\Big|^2
\label{eq7}
\end{equation}

In Equation \ref{eq7}, $\varepsilon$ is the overall residual error of a weighted combination of geometric error $\varepsilon_g$ and color error $\varepsilon_c$. The error is minimized using the Jacobean equation: $J^TJ \Delta x = -J \epsilon $. The overall Jacobean to find a converged solution is modeled as shown in Equation \ref{eq8}.

\begin{equation}
J_g = \lambda \omega_g (- n_r^T J_{3D} + (p_f - p_r )^T J_{rot})
\label{eq8}
\end{equation}

In Equation \ref{eq8}, $n_r=( x_n, y_n, z_n )^T$ and $p_r=( x_p, y_p, z_p )^T$. And geometric rows are computed using the following equations \ref{eq9} and \ref{eq10}.

\begin{equation}
J_{3D} = \begin{bmatrix} 1 & 0 & 0 & 0 & 2z_n & -2y_n \\ 0 & 1 & 0 & -2z_n & 0 & 2x_n \\ 0 & 0 & 1 & 2y_n & -2x_n & 0 \end{bmatrix} 
\label{eq9}
\end{equation}

\begin{equation}
J_{rot} = \begin{bmatrix} 0 & 0 & 0 & 0 & 2z_n & -2y_n \\ 0 & 0 & 0 & -2z_n & 0 & 2x_n \\ 0 & 0 & 0 & 2y_n & -2x_n & 0 \end{bmatrix}
\label{eq10}
\end{equation}

Further, to address the problem of drift because of loop closures, Henry et al have proposed a novel alignment procedure for global consistency in which they have divided the patch volumes into two sets: active patch volumes $S_{current}$ and inactive patch volumes $S_{old}$ where only path volumes of set $S_{current}$ take part into alignment and fusion. Only active patch volumes are used for computing loop closures along with re-rendering the inactive patch volumes. 

\subsection{RGB-D Visual Odometry}
Iterative closest point algorithm, though a popular algorithm for image registration, is prone to local minima as image registration is based on suboptimal initial point correspondences. The constraint is alleviated by the steps following registration like RANSAC and bundle adjustment but this increases computational cost while reducing sparse key points reduces computational cost at the expense of data loss. F. Steinbr$\ddot{u}$cker et al have proposed an energy-based framework for computing visual odometry from input data directly hence reducing computational cost \cite{steinbrucker2011real}. They have suggested a computation of rigid body motion that minimizez the least-squares error and maximizes photoconsistency a by finding a twist $\xi$ that minimizes the least-squares error as shown in Equation \ref{eq11}.

\begin{equation}
E(\xi) = \int\limits_\Omega [I(\omega_\xi (w,t_1), t_1) - I(x, t_0), t_0]^2 dx 
\label{eq11}
\end{equation}

They have proposed the neregy equation that minimizez the $\xi$ which is given as shown in Equation \ref{eq12}.

\begin{equation}
E_l (\xi) = \int\limits_\Omega \bigg( \dfrac{\delta I}{\delta t}+\Big( \nabla I.\dfrac{d\pi}{dG}.\dfrac{dG}{dg}.M_g\big|_{x,t_0}.\xi \Big)  dx  \bigg) 
\label{eq12}
\end{equation}

\begin{figure*}
	\centering
	\includegraphics[width=6in,height=1.3in]{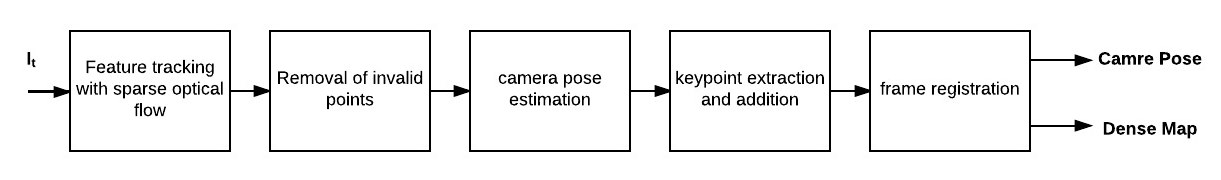}
	\caption{Visual Odometry approach for camera pose estimation and dense mapping}
\label{fig8}
\end{figure*}

An evaluation of the system over benchmark dataset by \cite{sturm2011towards} shows better accuracy compared to a general ICP algorithm that allows a better implementation of visual odometery computing algorithms on GPU.
These sollutions have also complimented the development of a surface mapping system for indoor environemnts in conjunction with visual odometery by Silva and Goncalves (\cite{silva2014visual}) that does not require a GPU for pose estimation. Unlike other feature detection through descriptor algorithms, Silva and Goncalves have adopted a short baseline optical flow tracker to track features for pose estimation. Features are tracked across consective image pairs $(I_{t-1}, I_t)$ via sparse optical flow (Figure \ref{fig8}) after initial Shi-Tomasi corner extraction. After the invalid points are removed, the resulting tracked points are added to robustly estimate the current camera pose $[R_t |t_t]$ using RANSAC. In the last setp the obtained RGB-D frame $I_t$ is registered with the previously estimated frame $I_{t-1}$ to obtain the 3D transform relative to the origin frame. 

The system has been evaluated for its localization accuracy based on Relative Position Error (RPE) compared with the RGB-SLAM system of Sturm et al using its publically available dataset \cite{sturm2011towards}. The evaluation shows competitive results of the approach proposed for volumetric reconstruction without using GPUs or any other expensive memory hardware. 

\subsection{Elastic Fusion}
\begin{figure*}
	\centering
	\includegraphics[width=7in,height=2in]{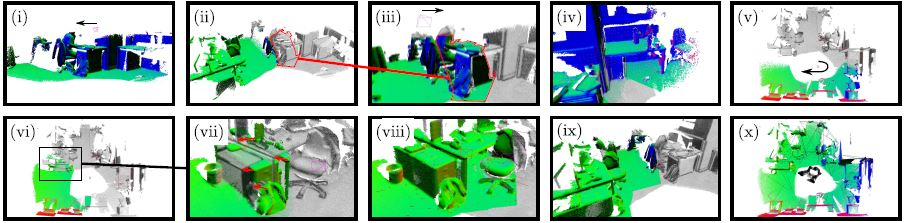}
	\caption{Main steps outlined for Elastic Fusion algorithm with a final global loop closure database containing sampled camera poses and underlying deformation graph}
\label{fig9}
\end{figure*}

All the volumetric approaches described above are mainly variants of kinetic fusion algorithm with modifications of different steps. While sparse methods focus on pose graph optimization for reconstruction Whelan et al have recently presented a map-centric approach \cite{whelan2015elasticfusion} for dense SLAM. Based on the offline dense surface reconstruction framework provided by \cite{zhou2013dense}, elastic fusion approach focuses on early and repeatedly loop closure optimizations. This allows for a non-rigid space formation of surface map (via model-to-model loop closures) through an embedded sparse deformation graph instead of a pose graph or any post-processing steps in online incremental manner. While local loop closure optimization keeps the data closer to map distribution global loop closures are computed to recover from arbitrary drift that ensure global consistency. \cite{turan2017non}
Figure \ref{fig9} shows the output of the elastic fusion algorithm. The algorithm is developed based on the point-based fusion algorithm developed by \cite{keller2013real} with a different approach for pose estimation. In camera tracking, the pose estimation is followed by joint optimization minimizing the joint cost function of geometric $E_{icp}$ and photometric $E_{rgb}$ pose estimation such that $E_{track} = E_{icp} + \omega_{rgb} E_{rgb}$. At each iteration Gauss-Newton non-linear least-square resulting in improved camera transformation as shown in Equation \ref{eq13}. 

\begin{equation}
T' = exp(\hat{\xi})T
\label{eq13}
\end{equation}

In Equation \ref{eq13}, $\hat{\xi} =  \begin{bmatrix} [\omega]_x & x \\ 0 0 0 & 0 \end{bmatrix} $. For local loop closures, instead of performing geometric frame-to-modal tracking via splatted rendering, elastic fusion performs photometric frame-to-modal tracking via full colored splatted rendering. Furthermore, surfels are marked active or inactive based on a set time window threshold $\delta_t$ and only active surfels are utilized for estimating camera poses and in depth map fusion. 

For global loop closures, \cite{whelan2016elasticfusion} have utilized the randomized fern encoding approach developed by \cite{glocker2015real} for adding predicted views and finding matches. In case of detected matches, the views are registered for computing globally consistent maps into a non-rigid deformation to obtain a global surface alignment \cite{turan2017non1}.
Compared with other state of the art systems including, Kinect fusion, DVO SLAM, and RGB-D SLAM, the results obtained from Elastic Fusion algorithm (Tables \ref{table1} and \ref{table2}) show robustness and better accuracy in terms of camera trajectory and surface estimation with accepted computational performance requirements without pose graph estimation or any other post-processing steps. The detailed results demonstrated in \cite{whelan2015elasticfusion} also show the efficacy of this approach in surface mapping at even larger scale than that of rooms or buildings (dense mapping for $t \rightarrow \infty$). 

\subsection{Bundle Fusion}
\begin{figure*}
	\centering
	\includegraphics[width=7in,height=2in]{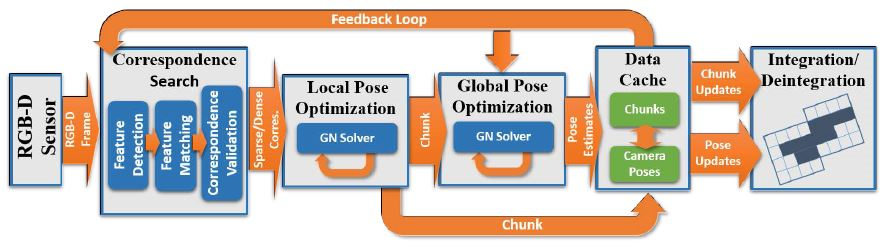}
	\caption{Bundle fusion pipeline for globally consistent 3D reconstruction with global pose optimization and on-the-fly surface re-integration – Courtesy: \cite{dai2017bundlefusion}}
\label{fig10}
\end{figure*}

To address the tracking gaps lying in systems discussed above, \cite{dai2017bundlefusion}, have recently proposed a novel robust pose estimation scheme in \cite{dai2017bundlefusion} that employs an efficient hierarchal approach for optimizing per frame to obtain a global set of camera poses. Taking into account all the RGB-D history, bundle fusion approach sets its core to a novel two-stage global pose optimization strategy that results in an online globally-consistent 3D mapping. First, a sparse-then-dense global pose optimization is performed to obtain global alignment (Figure \ref{fig10}). A live RGB-D stream from the sensor is input to the system followed by an optimal rigid camera transforms by 3D correspondences between the frames. Correspondence filtering incorporates both geometric and photometric optimization to minimize outliers through key point correspondence filter, and surface area filter. 

Dai et al have followed hierarchical optimization approach in which input scene sequences are divided into small chunks of consecutive frames and local alignments are achieved within the chunks followed by global alignment in the next step. After performing a hierarchal optimization and pose estimation, dynamic 3D reconstruction is performed while continuous monitoring an updating consecutive poses through integration and disintegration of frame which helps in fixing the problem of accumulated drift and dead reckoning in feature-less regions. Same as the approach of \cite{silva2014visual}, for a parameter vector $X=(R_o, t_o, \cdots , R_{|S|}, t_{|S|})^T$ for $|S|$ frames, the pose alignment based on an energy optimization approach is achieved as given in Equation \ref{eq14}.

\begin{equation}
E_{align}(X) = \omega_{sparse} E_{sparse}(X) = \omega_{dense}E_{dense}(X)
\label{eq14}
\end{equation}

In Equation \ref{eq14}, $ω_{sparse}$ and $ω_{dense}$ are weights assigned to sparse and dense matching terms and $E_{sparse} (X)$ and $E_{dense} (X)$ are the sparse and dense matching terms respectively, such that: $E_{sparse} (X) = \Sigma_{i=1}^|S| \Sigma_{j=1}^|S| \Sigma_{(k,l) \in C(i,j)} \Big|\big| T_iP_{i,k} - T_j P_{j,l} \big|\Big|^2$ with $T_i$ being rigid camera transformation, $P_{i,k}$ the $k^{th}$ detected feature point in $i^{th}$ frame, and $C(i,j)$ being the set of pairwise correspondences between the $i^{th}$ and $j^{th}$ frame. And $E_{dense} (T) = \omega_{photo} E_{photo} (T) + \omega_{geo} E_{geo} (T)$ with $\omega_{photo}$ and $\omega_{geo}$ being the weights of photometric and geometric term and $E_{photo} (T)$ and $E_{geo} (T)$ being terms for calculating photometric and geometric alignment respectively. 
Pose optimization is performed at every frame and reconstruction is updated accordingly along free-camera paths instead of temporal coherence. To make global pose alignment real-time and tractable a hierarchal local-to-global pose optimization is performed ensuring increased scalability for larger scene reconstruction. Dense surface reconstruction is achieved via volumetric fusion reconstruction pipeline proposed by Nießner et al with an added novelity of symmetric on-the-fly RGB-D farmes registration. 

\begin{figure*}
	\centering
	\includegraphics{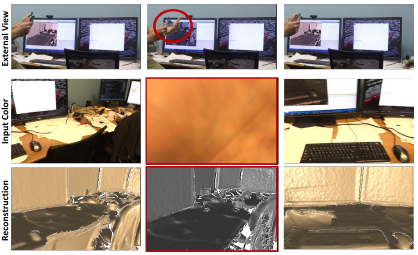}
	\caption{Results showing Bundle Fusion's effectiveness in detecting grey overlay and recovering from tracking failures}
\label{fig11}
\end{figure*}

Where all frame-to-model tracking algorithms (\cite{newcombe2011kinectfusion}, \cite{chen2013scalable}, \cite{niessner2013real}) using ICP for closure detection face the problem of accumulated drift, bundle fusion outperforms them in terms of recovery from tracking failure (Figure \ref{fig11}) and reducing geometeric drift through an implicit global pose otimization approach to handle loop closures. 

\section{Milestones about Kinect and bundle fusion and its derivations as a survey and history summarization}
\subsection{Kinetic fusion}
Kinetic Fusion refers to the 3D object scanning and model development resource that utilizes a Kinetic for Windows based operating system sensor. The user is able to paint a scene using the Kinetic camera while simultaneously viewing and interacting with an extensive 3D model of the scenery. Kinetic fusion is able to be operated within interactive rates with GPU support and is also able to operate on non-interactive rating on varied hardware. 
However, it should be noted that operating at non-interactive rates can permit substantial volume reconstructions. Kinetic Fusion is able to process data by using either DirectX 11 GPU compatibility with C++ AMP with the alternative being on the CPU which is attained by placement of the reconstruction processor format in the course of the reconstruction volume development. The CPU processor is well suited for applications of offline processing which is restricted to DirectX 11 GPUs that permit for real time and interactive frame rating in the process of reconstruction. 

\begin{figure*}
	\centering
	\includegraphics[width=7in,height=3in]{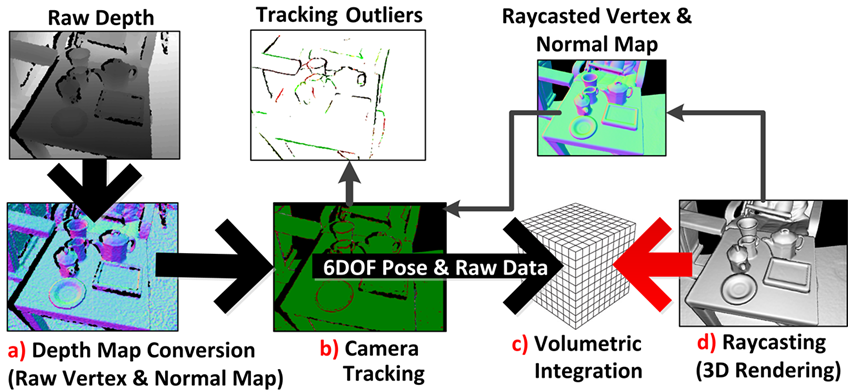}
	\caption{Graphical chart representation of Kinetic fusion processing pipeline entailing multiple stages from raw depth to attain 3D reconstruction}
\label{fig12}
\end{figure*}

Recommended hardware entails Desktop computers with GHz or better utilizing a multi-core processor as well as a graphics card an additional dedicated on-board memory. It has also been tested on high end platforms of NVidia GeForce GTX680 as well as AMD Radeon HD 70. 
The capability also exists of Kinetic Fusion to be operated on a laptop grade of DirectX11 hardware even as it normally operates on a significantly reduced performance in comparison of desktop grade hardware. In overall, the objective seeks to process a similar frame rate at 30fps as the Kinetic sensor to allow for the extensive robustness of tracking camera poses.

\subsection{Bundle fusion}
Bundle Fusion on the other hand allows for real-time, high grade 3D scanning of extensive scale scenery as a major consideration to mixture of reality and robotic applications. However, scalability introduces the challenge of drifting with estimation of poses and the introduction of significant errors in the accumulation models. This translates into approaches normally requiring extended hours for offline processing of international correct model errors. 
Current online approaches has provide compelling outcomes even as they are limited from several issues. The first issue is the requirement of minutes to undertake online correction in the prevention of actual time usage. The second issue is the brittle frame to frame model pose estimation as an outcome in several tracking shortcomings. The third and final issue is their support is restricted to unstructured point based representation that is able to limit the quality of scanning and application. 

\begin{figure*}
	\centering
	\includegraphics[width=7in]{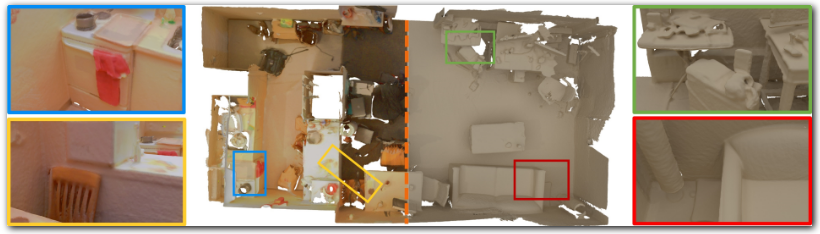}
	\caption{Graphical representation of real time, 3D scanning and high quality large scaling is vital to attaining a mixed reality which can be applied to robotic applications}
\label{fig13}
\end{figure*}

This can be addressed systematically to concerns where a novel, actual time and end to end reconstruction frame working. Within its core is the estimation of robust possess based on strategies that optimize the per frame rating for a global collection of camera posing with consideration of the entire history of RGB-D input using efficient hierarchical methods. The heavy reliance on temporal tracking is eliminated and persistently localized to the global optimization of frames. 
The contribution is a parallel optimization framework that is able to apply correspondence on the basis of spatial feature and heavy geometric and photometric matching. The approach used is an estimation of global optimization which is to say bundle adjustment that places real time and support for robust tracking using gross tracking failure recovery which is to say re-localization and re-estimation of the 3D model in actual time for ensuring the global consistency all in entire single framework. This approach outperforms that modern online systems using quality on par to offline approaches even with unprecedented speeding and scan completion.

\section{Algorithms Characterization and Analysis}
The performance of the RGB-D SLAM and Kinect Fusion algorithm discussed above has been summarized in table 1 below in terms of absolute trajectory room mean square metric (ATE RMSE) on two datasets: of \cite{handa2014benchmark} and of \cite{sturm2012benchmark}. Comparison is done based on scene tracking, offline or online registration performance, camera revisiting the previous map, computing loop closures, and surface reconstruction. 
The algorithms are usually in par with or are better than others in some features and lack in other and are well suited for various 3D reconstruction applications based on requirements and level of complexity. But some important points to note here are the computing performance of these algorithms to be implemented on commercial scale and more importantly efficacy of these algorithms in performing a live, real-time, large scale, and drift-free dense mapping of the scenes. 

\begin{table}[!t]
\renewcommand{\arraystretch}{1.3}
\caption{Comparison of trajectory estimation in terms of ATE RMSE results of state-of-the-art algorithms on the evaluated synthetic ICL-NUIM datasets of \cite{handa2014benchmark} Source: (Manipulator and object tracking for in-hand 3D object modeling) \cite{krainin2011manipulator}}
\label{table1}
\centering
\begin{tabular}{|c|c|c|c|c|}
\hline
System & Kt0 & Kt1 & Kt2 & Kt3\\
\hline
DVO SlAM & 0.104m & 0.029m & 0.191m & 0.152m\\
\hline
RGB-D SLAM & 0.026m & 0.008m & 0.018m & 0.433m\\
\hline
MRSMap & 0.204m & 0.228m & 0.189m & 1.090m\\
\hline
Voxel Hashing & 0.014m & 0.004m & 0.018m & 0.120m\\
\hline
Kintinuous & 0.072m & 0.005m & 0.010m & 0.355m\\
\hline
Frame-to-model & 0.497m & 0.009m & 0.020m & 0.243m\\
\hline
Elastic Fusion & 0.009m & 0.009m & 0.014m & 0.106m\\
\hline
Bundle Fusion & 0.006m & 0.004m & 0.006m & 0.011m\\
\hline
\end{tabular}
\end{table}

\begin{table}[!t]
\renewcommand{\arraystretch}{1.3}
\caption{Comparison of trajectory estimation in terms of ATE RMSE results of state-of-the-art algorithms on the evaluated synthetic TUM RGB-D dataset. Source: (Manipulator and object tracking for in-hand 3D object modeling) \cite{krainin2011manipulator}}
\label{table2}
\centering
\begin{tabular}{|c|c|c|c|c|}
\hline
System & fri/desk & fr2/xyz & fr3/office & fr3/nst\\
\hline
DVO SlAM & 0.021m & 0.018m & 0.035m & 0.018m\\
\hline
RGB-D SLAM & 0.02m & 0.008m & 0.032m & 0.017m\\
\hline
MRSMap & 0.043m & 0.020m & 0.042m & 2.018m\\
\hline
Voxel Hashing & 0.023m & 0.022m & 0.023m & 0.087m\\
\hline
Kintinuous & 0.037m & 0.029m & 0.030m & 0.031m\\
\hline
Frame-to-model & 0.022m & 0.014m & 0.025m & 0.027m\\
\hline
Elastic Fusion & 0.020m & 0.011m & 0.017m & 0.016m\\
\hline
Bundle Fusion & 0.016m & 0.011m & 0.022m & 0.012m\\
\hline
\end{tabular}
\end{table}

The systems developed either perform implicit registration of poses and suffer from the issues of scalability because of limitations of uniform grid as in the case of Kinect Fusion \cite{newcombe2011kinectfusion}; or in case of more efficient volumetric fusion strategies (\cite{chen2013scalable} \cite{niessner2013real} \cite{whelan2012robust} \cite{roth2012moving}) face increased drift because of pose errors in pose estimation. The approaches performing globally consistent mapping suffer from a lack of live real-time reconstruction because of offline processing and larger memory and processor requirements (\cite{zhou2013dense} \cite{whelan2012kintinuous}). 
As pointed out by \cite{dai2017bundlefusion}, a comprehensive and state-of-the-art system is required to meet the requirements of high-quality surface modeling that is able to model continuous surfaces while meeting the requirements of robust camera tracking, real-time rates, on-the-fly modal updates, scalability, and global model consistency. Although bundle fusion seems to address these issues in future providing a complete robust tracking and mapping system but the system still suffers from misalignments and memory issues. 
Thus, for the environments that are too challenging or the robot dynamics are complex, the SLAM algorithms developed for real-time tracking and mapping may fail and there is still need for research in this arena \cite{turan2018sparse}. The SLAM algorithms are considered to have been entered into the third era of their development which Cadena et al called the robust-perception age \cite{cadena2016past}. The requirements of the robust-perception age the SLAM algorithms are currently in are pointed out by \cite{cadena2016past} as: robust performance of the SLAM system that allows fail-safe performance of the system with low failure over an extended period of time with self-tuning characteristics unlike current algorithms that require manual tuning; high-level understanding of the environment by the system, resource awareness in which the system can adapt to available sensing and computational resources; and task-driven perception of the system in which system is able to discard irrelevant data by itself allowing adaptive representation of maps whose complexity may vary depending on the type of task being performed. 
The approaches discussed here all use the standard front-end sensing and back-end processing approach in which the front-end senses and gathers the data while the back-end does the processing of that data input by the front-end \cite{cadena2016past}. Algorithms relying on this approach are always prone to loop closure problem as the input of wrong loop closures to the back-end processing is unavoidable which further degrades the quality of the maps generated. Although the systems like in \cite{henry2012rgb} \cite{curless1996volumetric} \cite{glocker2015real} have adopted novel strategies to avoid the problem of wrong loop closures but the problem still requires more focus in terms of improving the robustness of the systems. a new line of research has been initiated specifically to deal with this problem as identified in \cite{sunderhauf2012towards} and \cite{latif2013robust}. Another problem of such algorithms is that they are prone to outliers \cite{cadena2016past} and this is where the fail-safe or the automatic recovery of the system comes as the key requirement for a robust system. Bundle fusion \cite{glocker2015real} shows a much better metric in this case than other SLAM systems discussed here. 
In terms of scalability, most of the systems discussed in the paper show an effective performance for small indoor environment but have made improved scalability their future task showing there is much work to be done for systems’ scalability to map larger outdoor environments over an extended period of time. The approaches mentioned use the camera re-visiting approach using iterative linear solvers to register more scenes. This approach requires much larger memory for the mapping of larger environment \cite{sturm2012benchmark} and hence put a constraint on the scalability of the systems due to larger memory and computing requirements. However, the systems like the one developed by \cite{huang2017visual} that has incorporated GPU-based processing for a scalable mapping and such shown in \cite{kainz2012omnikinect} and \cite{chen2013scalable} show some prolific prospects for better scalability of the SLAM systems. 
The use of Kinect sensor for simultaneous localization and mapping with considerable outcomes predicts an opportunity of better SLAM systems with improved novel sensors that allow active sensing and real-time scalable mapping by using such systems. New tools and technologies are needed to be integrated to obtain complete robust, fail-safe, self-tuning SLAM systems that would be able to predict, update, remember, or discard, the information according the requirements of the tasks in hand by utilizing the resources that are available and by adapting to the environmental requirements.
Several applications of relevant in robotics as well as computer vision need the capability of rapid acquisition of 3D models of the environment and the estimation of the camera pose in respect to the model. A robot for instance, requires ascertaining its location for navigation between places. This challenge is classical and difficult since camera localization in needs 3D models that is turn call for the camera pose according to \cite{niessner2013real}. As such, the camera trajectory and 3D model require estimation at the similar timing. Introducing Kinetic fusion platform avails for colored imagery and dense depth mapping for complete video frame rating. 
 It also permits for the creation of a novel method to SLAM integrates the scaling information of 3D depth sensing with the benefits for visual features to creating dense 3D setting representations. The estimation of the trajectory is segmented into a frontal end and a back end while the frontal extraction spatial relation among the specific observations provides for a back end optimization that places such observations in a so defined pose graph respecting non-linear error functions as interpreted by \cite{agarwal2010reconstructing}.
 The frontal end utilizes visual imagery of the RGB-D sensor for detection of the key points as well as extracts descriptors. These can be matched to past extraction descriptors as well as the relative transformation among the sensor pose based on computation with use of the RANSAC by \cite{glocker2015real}.
Our analysis is based on the novel RGB-D SLAM system of visual odometry and information filter extension that does not need any other sensors or odometry. This is different to the approaches of graph optimization which is increasingly suitable for online applicability. The visual dead reckoning algorithm founded on the visual residuals is formulated which is applied in estimation of motion control input as presented by \cite{niessner2013real}. This is augmented with the utilization of the novel descriptor known a binary robust appearance and normalized descriptors (BRAND) for the extraction of features from the RGB-D frame and utilizing them in the form of landmarks according to \cite{karan2015calibration}.
 In addition, with consideration of the 3D positions and landmark BRAND descriptors, we shall use an observation model that limits explicit data relation between the observations and mapping with the use of marginalization observation possibility over entire likely relations. 
Experimental validity is availed in comparison with the proposed RGB-D SLAM algorithm using mere RGB-D visual odometry as well as graphing as provided by the dataset. The analytical results of the dataset reveal the self localization is broadly held in recognition as one of the most elementary challenges for robotics autonomy in regard to navigation according to \cite{curless1996volumetric}. This assignment can be undertaken well at the point when the environment is defined as priori even the mapping is not availed beforehand. Therefore, robot localization becomes highly challenging.
This can be attributed to the inadequate setting information of the movement of the robot in or the excessive expense of manually constructing a map based on objective. In this instance, the robot needs to simultaneously formulate a map of the environment followed by self-localization in it. This challenge defined as simultaneous localization and map construction is extensively examined. 
The SLAM solutions challenge presented by far differs majorly for the setting description adopted and the employed estimation technique. The two main estimation methods we used for analysis are filter and graph based SLAM. Filter based SLAM entails estimation of the posterior using the means of the Bayes' rule shown in Equation \ref{eq15}. 

\begin{equation}
p(\xi_t, m|z_{1:t}, u_{1:t})) 
\label{eq15}
\end{equation}
 
In Equation \ref{eq15}, $\xi_t$ is the robot pose at time t, m denotes the map, $z_{1:t}$ as the observation sequence as well as $u_{1:t}$ as the odometry information. Also known as online SLAM, it uses an incremental past measurement and control alternative which are neglected upon processing. In accordance to varied approaches of tacking the posterior probability, there are several filter based techniques such as the extended Kalman filter (EKF), the extended information filter (EIF) as well as the particle filter method (PF).
 Rather than estimation of the single post $\xi_t$, within filter based SLAM, the graph based estimation of the full trajectory $\xi_{1:t}$ as well as the map denoted by m is for the entire information observed by \cite{zhou2013dense}. Even as this approach is held in consideration as being time consuming and unable to satisfy actual time needs, however by techniques of efficient solving, graph based SLAM avails the more attention.
The initial analysis on the SLAM challenge places emphasis on the two dimensional setting such that can be normally applied in mobile robotics. Of recent, \cite{krainin2011manipulator} propose varied 3D SLAM algorithms have provided supports for 6-DOF (degree of freedom) estimation of pose such that the employed SLAM technique in varied platforms for instance quad rotors among others. Earlier 3D SLAM studies, costly sensors such as 2D and 3D-laser range finders were major applied. \cite{2018arXiv180307608S}
However, of recent, with the introduction of low cost Kinetic style sensors known a RGB-D cameras, they provide color imagery and depth data in a concurrent manner that is known as RGB-D SLAM  \cite{turan2018deep}. For the greater part of our robotics SLAM, this was undertaken with a sensor that avails 2D scenery with the major reasoning being that in order to attain high quality 3D data, the cost is high. It is at this point that the cheap Kinetic technology provides immense interest in the capture and reconstruction of 3D environments using a RGB-D sensor that is moveable. It avails dense, increasingly resolution depth information at a lower cost and scope on the basis of data by \cite{sturm2012benchmark}. 
We formulate SLAM application with the use of Kinetic and bundle adjustment framework for integration of the iterative closest point using visualized feature matching. Our research, our graph optimization uses a g2o also known as the general optimization framework to attain global alignment according to \cite{zhou2013dense}. This means we adopt the iterative closest point for pairwise alignment among the sequential framing and recovering of the rigid transformation within point clouds. The accuracy alignment of the iterative closest point is significantly dependant on the content of the scene. We therefore used color feature descriptors for improvement of our depth data correspondence.
We therefore propose a RGB-D slam approach that manages low dynamic scenarios with the use of pose graph structures where grouping of nodes is based on covariant values. Any constraints that are falsified are pruned on the basis of error metrics associated to the node factions according to \cite{keller2013real}. Our study therefore examines highly efficient pose graph optimization for instance tree based network optimizer (TORO). 
Improvement of this algorithm when combined with features from accelerated segment test (FAST) a well a Calonder descriptors is able to provide an estimation of the pose with use of the re-projection error random sample consensus (RE-RANSAC) methodology for frame to frame based alignment as well as incorporation of ICP constraints into sparse bundle adjustment (SBA) for global optimization.
 The algorithm core known as RGB-D ICP entails a noble iterative closest point variant that takes use of the extensive information within RGB-D data. Our methodology is highly efficient in complex indoor environments as its trajectory algorithmic estimation is based on the integration into singular global procedures that are not reliant on intermediate level features which has also been observed in works by \cite{dai2017bundlefusion}. 
More so, the use of accurate pose measurement with techniques of localization and a compact photometric environment model is attained. In the correct rigid body motion of a handheld RGB-D camera, the energy based approach is the estimation. 
This when combined with the technology of odometry sensor RGB-D for automated flight experimental analysis creates the possibility of planning complex 3D paths within a cluttered setting. This presents us with a novel GPU implementation on the foundation of an RGB-D visual odometry algorithm using the 6-DOF camera odometry estimation methodology for tracking and integrating RGB color information into the reconstruction process of Kinetic Fusion to permit a high quality mapping. 
The analysis results reveal there is no necessity for the application of key frames and the outcomes of real-time colored volumetric surface reconstructions shows the several RGB-D SLAM techniques are restricted to geometrically structured environments. At this point we can propose a switched based algorithm with heuristic selection between RGB-D bundle adjustments on the basis of localized map building. Such maps are developed by the application of sparse bundle adjustments on an inclusion of two-step re-projection of RANSAC and ICP approach.
 With a heuristic switching algorithm, we deal with multiple failure modes related with the RGB-D-BA bundle adjustment. The map linkage mechanism greatly lowers the computational expenses such that this algorithm holds immense benefit for application in a large scale setting. For evaluation of the system, we shall utilize the RGB-D benchmark that avails a Kinetic sequence dataset using synchronized ground truth. In addition, the benchmark avails an evaluation resource based on computation of the root mean square error when provided with an estimated trajectory.
 In evaluating, we select the Freiburg 1 dataset comprising of nine sequences in placement with a normal indoor environment. Two of these sequences hold very simplified motions such that the outcome of the sequences reveals the capabilities of this technique in the best case.

\begin{table*}[!t]
\renewcommand{\arraystretch}{1.3}
\caption{System evaluation on a large set of sequences from RGB-D SLAM dataset with a 9.7cm and 3.95 degrees accuracy and a required approximation of 0.35 seconds for time per image processing. Source: (RGB-D mapping: Using Kinect-style depth cameras for dense 3D modeling of indoor environments) \cite{henry2012rgb}}
\label{table3}
\centering
\begin{tabular}{|p{1cm}|p{1cm}|p{1cm}|p{2cm}|p{2cm}|p{1cm}|p{1cm}|p{1cm}|p{1cm}|p{1cm}|}
\hline
Sequence Name & Length & Duration & Avg. Angular Velocity & Avg. Transl. Velocity & Frames & Total Runtime & g20 Runtime & Transl. RMSE & Rot. RMSE\\
\hline
FR1 360 & 5.82m & 28.69s & 41.60deg/s & 0.21m/s & 745 & 145s & 0.66s & 0.103m & 3.41deg\\
\hline
FR1 desk2 & 10.16m & 24.86s & 29.31deg/s & 0.43m/s & 614 & 176s & 0.68s & 0.102m & 3.81deg\\
\hline
FR1 desk & 9.26m & 23.40s & 23.33deg/s & 0.41m/s & 575 & 199s & 1.31s & 0.049m & 2.43deg\\
\hline
FR1 floor & 12.57m & 49.87s & 15.07deg/s & 0.26m/s & 1214 & 488s & 3.93s & 0.055m & 2.35deg\\
\hline
FR1 plant & 14.80m & 41.53s & 27.89deg/s & 0.37m/s & 1112 & 424s & 1.28s & 0.142m & 6.34deg\\
\hline
FR1 room & 15.99m & 48.90s & 29.88deg/s & 0.33m/s & 1332 & 423s & 1.56s & 0.219m & 9.04deg\\
\hline
FR1 rpy & 1.66m & 27.67s & 50.15deg/s & 0.06m/s & 687 & 243s & 10.26s & 0.042m & 2.50deg\\
\hline
FR1 teddy & 15.71m & 50.82s & 21.32deg/s & 0.32m/s & 1395 & 556s & 1.72s & 0.138m & 4.75deg\\
\hline
FR1 xyz & 7.11m & 30.09s & 8.92deg/s & 0.24m/s & 788 & 365s & 40.09s & 0.021m & 0.90deg\\
\hline
\end{tabular}
\end{table*}

However, outcomes such as this can normally be attained with the careful movement of the sensor in an indoor environment. For example, in the course of manual mapping recording before deploying a robot. The other datasets are increasingly complex since they entail coverage of larger sections as well as unrestricted camera motions. 
The results of the accuracy of the SLAM system and evaluation are dependent on the accuracy of the selected feature detector and sensor frame rating \cite{turan2017deep1}. The second step is the investigation of the influence of multiple parameter4s on the system runtime. As observed on the entire nine sequences of the system in the FR1 table, the average camera velocities fall in the range of 9 to 42 degrees as well as 0.06 to 0.43 meters per second.

\subsection{Feature detection and descriptor extraction}
 The method for detecting frames as well as extracting descriptors is increasingly apparent when specific to incoming image frames. The table below is a representation of the time comparison required for distinct feature types founded on early descriptives. From the tabulated information, ORB holds a faster speed compared to SURF and SIFTGPU according to single order magnitudes. However, the results based on revelations of increased errors in sequences of two or nine re-aimed angle at almost of 3.5x speeds when compared to SURF implementation that is non-parallel.

\begin{table}[!t]
\renewcommand{\arraystretch}{1.3}
\caption{Feature runtime analysis with respect to a selected interest point detector and feature descriptor. Source: (RGB-D mapping: Using Kinect-style depth cameras for dense 3D modeling of indoor environments) \cite{henry2012rgb}}
\label{table4}
\centering
\begin{tabular}{|p{1cm}|p{3cm}|p{3cm}|}
\hline
Type & Avg. Count - Std. Dev & Runtime Detection + Extraction Avg. - Std. Dev.\\
\hline
SURF & 1733 - 153 & 0.34s + 0.34s\\
\hline
ORB & 1117 - 558 & 0.018s + 0.0086s\\
\hline
SIFTGPU & 1918 - 599 & 0.19s\\
\hline
\end{tabular}
\end{table}

Matching of features and estimating motion needs computation of single instances per frames in case the present frame is limited to matched single predecessors such that the output camera trajectory is rapidly increased with accumulation of errors when measured against time. Multiple frames feature matching is costly to compute more so since no assumption is held of the availed odometry information being restricted to the likelihood of closures within loops. The system can also be accurate of trajectories that are extended with augmented information concerning pairwise relative transformation that makes estimating of the trajectory highly robust to errors when estimated according to pose. However, this method is highly linked and requires increased optimization time. Therefore, we can understand that match current factors to past 20 frames provide a satisfactory compromise.

\begin{table}[!t]
\renewcommand{\arraystretch}{1.3}
\caption{Runtime analysis of pair wise frame registration. Source: (RGB-D mapping: Using Kinect-style depth cameras for dense 3D modeling of indoor environments) \cite{henry2012rgb}}
\label{table5}
\centering
\begin{tabular}{|p{1cm}|p{2cm}|}
\hline
Matcher & Runtime (Avg. - Std.  Dev.)\\
\hline
FLANN & 0.203s - 0.078s\\
\hline
Brute Force & 0.386s - 0120s\\
\hline
\end{tabular}
\end{table}

This is reflected in the tabulated information above in form of the mean run time in estimating and matching motion which is also a revelation of FLANN reducing the time required for registering frames by a factor of two. Pose graph optimization. Optimization of minimal graphs is faster to implement in the real time as well as for entire frames with lengthy sequences and dense connections with increases in time optimization. However, in case estimating motion is reliable, the overall method can be implemented at all levels according to findings by \cite{whelan2015real}. 
In full sequences, graph optimization ratio when compared to the general time is lower than 6 percent. The novel method to visualization of SLAM algorithms with the use of RGB-D sensors means the method will entail the methodologies that entail extracting visual key points from colored images and applying depth images to localize. It is at this point that we can use RANSAC for robust estimation of changes between RGB-D frames and pose graph optimization using non-linear techniques. At the end, volumetric 3D maps of the setting can be used to localize the navigation of the robot and plan its path.

\section{Conclusion}
All the reconstruction approaches developed till now contribute towards a specific modification in 3D reconstruction algorithms for efficient camera tracking and surface mapping. But the systems developed till now still lack one crucial element: deployment of a complete robust and accurate real-time tracking and mapping system to be used for robotic navigation, scanning and mapping. RGB-D sensors have opened a new field of opportunities to handle this challenge and paved a way for improved (if not completely solved) SLAM systems in future. Prototype systems developed do show promising efficacy towards a universal real-time 3D reconstruction system implementation in future.

\bibliographystyle{IEEEtran}
\bibliography{mybibfile}

\end{document}